\documentclass[runningheads]{llncs}
\usepackage[T1]{fontenc}
\usepackage{graphicx}
\usepackage{amsmath}
\usepackage{amssymb}
\usepackage{booktabs}
\usepackage{array}
\usepackage{multirow}
\usepackage{color}
\usepackage{colortbl}
\usepackage{framed}
\usepackage{bm}
\usepackage{bbm}
\usepackage{xspace}
\usepackage{enumitem}
\usepackage{cite}
\usepackage{xparse}
\usepackage{xcolor}
\usepackage{tikz}
\usetikzlibrary{backgrounds}
\usetikzlibrary{arrows,shapes}
\usetikzlibrary{tikzmark}
\usetikzlibrary{calc}
\usepackage{algorithm}
\usepackage{lipsum}
\usepackage{listings}
\usepackage{url}
\usepackage{algorithmic}
\usepackage{pifont}
\usepackage{bbding}
\usepackage{mathtools}
\usepackage{cleveref}

\definecolor{Gray}{gray}{0.92}
\definecolor{LightCyan}{rgb}{0.88,0.95,1}
\definecolor{blond}{rgb}{0.98, 0.94, 0.75}

\def \ie {\emph{i.e.}}
\def \eg {\emph{e.g.}}
\def \etal {\emph{et al.}}

\newcommand{\tit}[1]{\smallbreak\noindent\textbf{#1.}}
\newcommand{\tinytit}[1]{\noindent\textbf{#1.}}

\newcommand{\ours}{Self-Cap\xspace}

\begin{document}
\title{Fluent and Accurate Image Captioning\\with a Self-Trained Reward Model}

%
\titlerunning{Fluent and Accurate Image Captioning with a Self-Trained Reward Model}
%
\author{Nicholas Moratelli \and 
Marcella Cornia \and 
Lorenzo Baraldi \and 
Rita Cucchiara}
\authorrunning{N. Moratelli et al.}
%
\institute{University of Modena and Reggio Emilia, Italy\\
\email{\{name.surname\}@unimore.it} 
}
\maketitle              
\sloppy
\begin{abstract}
Fine-tuning image captioning models with hand-crafted rewards like the CIDEr metric has been a classical strategy for promoting caption quality at the sequence level. This approach, however, is known to limit descriptiveness and semantic richness and tends to drive the model towards the style of ground-truth sentences, thus losing detail and specificity. On the contrary, recent attempts to employ image-text models like CLIP as reward have led to grammatically incorrect and repetitive captions. In this paper, we propose \ours, a captioning approach that relies on a learnable reward model based on self-generated negatives that can discriminate captions based on their consistency with the image. Specifically, our discriminator is a fine-tuned contrastive image-text model trained to promote caption correctness while avoiding the aberrations that typically happen when training with a CLIP-based reward. To this end, our discriminator directly incorporates negative samples from a frozen captioner, which significantly improves the quality and richness of the generated captions but also reduces the fine-tuning time in comparison to using the CIDEr score as the sole metric for optimization. Experimental results demonstrate the effectiveness of our training strategy on both standard and zero-shot image captioning datasets.
\end{abstract}

\section{Introduction}
\label{sec:intro}
The image captioning task involves a step-by-step generation of textual descriptions, where each word is produced incrementally. During this process, contextual information is taken into account by leveraging the previously generated words while also incorporating the semantic information derived from the visual features of the input image. Over the years, researchers have made remarkable progress in developing image captioning architectures in such a way that the model strives to produce captions that effectively capture the salient aspects of the image while maintaining linguistic fluency and relevance. In the initial stages, traditional training of early architectures involved minimizing the standard cross-entropy loss. Subsequent advancements introduced reinforcement learning techniques based on policy gradient methods, as proposed by~\cite{ranzato2016sequence,liu2017improved}. Similarly, the most adopted paradigm employs SCST (Self-Critical Sequence-Training)~\cite{rennie2017self}, which has demonstrated notable improvements in achieving state-of-the-art results through the optimization of the CIDEr metric~\cite{vedantam2015cider}.

Despite substantial progress, the capability to generate ``human-like'' descriptions remains a challenge. Recently, there has been an exploration of the large-scale CLIP model~\cite{radford2021learning} for evaluating image captioning performance. This led to the development of the CLIP-Score~\cite{hessel2021clipscore}, which demonstrated a considerable correlation with human judgment, thereby highlighting its effectiveness as an evaluation metric. Following this direction, other evaluation metrics based on the CLIP model have been proposed~\cite{sarto2023positive,wada2024polos,sarto2024bridge}. Among them, PAC-Score~\cite{sarto2023positive} stands out for its greater correlation with human evaluations, obtained thanks to a positive-augmented fine-tuning strategy that has converted the CLIP embedding space towards the style of COCO captions~\cite{lin2014microsoft}. When employed as a reward for a captioning model, these metrics exhibit impressive ability to generate semantically rich sentences. Nonetheless, they also lead to significantly longer captions that may often contain word repetitions and grammatical errors and tend to overlook the proper word order in captions, which is an essential prerequisite in text generation.

To address these issues, we propose a novel approach based on SCST, wherein the image captioning model learns to generate captions by iteratively refining its output through a self-evaluation mechanism. Our strategy encompasses two key steps. First, we conduct a fine-tuning process for a caption discriminator using a self-supervised methodology inspired by CLIP. Specifically, alongside the usual positive image-caption pairs, we introduce a set of negative texts generated by the captioning model fine-tuned with the original CLIP-S and PAC-S as reward. The overall goal is to create a self-supervised environment that improves the correlation with human judgment, preserves syntactic accuracy, and allows the model to learn from its errors. As a second step, we integrate this discriminator as the reward used to fine-tune a captioning model, further enhancing its ability to generate high-quality and semantically richer captions.

We assess the effectiveness of the proposed approach by conducting several experiments on the COCO dataset~\cite{lin2014microsoft}, thereby showcasing its robust performance across a range of different backbones. To enhance the comprehensiveness of our analysis and validate the zero-shot capability of our approach, we expand our investigations to include out-of-domain experiments conducted on additional datasets like CC3M~\cite{sharma2018conceptual}, nocaps~\cite{agrawal2019nocaps}, and VizWiz~\cite{gurari2020captioning}, providing insights into its potential applicability in various real-world scenarios. 

\section{Related Work}
\label{sec:related}
\tinytit{Standard image captioning architectures}
Early captioning architectures initially involved filling in predefined templates after identifying relevant objects within the image~\cite{socher2010connecting,yao2010i2t}. Notable advancements in this field led to the adoption of CNNs for encoding images, traditionally employed in several Computer Vision tasks~\cite{pollastri2021deep,pollastri2021confidence,2018IRCDL}, followed by RNNs to describe the encoded visual information into natural language~\cite{vinyals2015show,rennie2017self,karpathy2015deep}. This approach was further refined with the incorporation of attention mechanisms~\cite{lu2017knowing,xu2015show}, which facilitated a shift towards enhancing the generation by focusing on key regions in the image~\cite{anderson2018bottom}, eventually enriched with spatial and semantic graphs~\cite{yao2018exploring,yang2019auto}. Currently, in addition to shifting towards Transformer-based architectures~\cite{cornia2020meshed,huang2019attention,cornia2022explaining}, a dominant strategy involves leveraging visual features from comprehensive cross-modal architectures like CLIP~\cite{shen2022much}. In this context, several directions have been explored, such as defining memory concepts to gather information from other samples~\cite{cornia2020meshed,barraco2023little} or integrating external knowledge into the architecture~\cite{li2022comprehending}. More recently, the advent of large scale models like LLMs and multimodal LLMs~\cite{touvron2023llama,vicuna2023,caffagni2024r,caffagni2024wiki} as significantly changed the landscape of image description leading to generated captions with increased descriptive capabilities~\cite{bucciarelli2024personalizing,dong2024benchmarking,li2024if}.

\tit{Training strategies}
While initial captioning models were trained with a standard cross-entropy loss~\cite{vinyals2015show,karpathy2015deep,xu2015show}, literature in this field soon turned towards the use of reinforcement learning paradigms. This strategy entails conceptualizing the models as agents, with the primary goal of maximizing the expected reward. On this line, notable advancements have been made by adopting a reinforcement learning strategy defining the reward as non-differentiable metrics~\cite{ranzato2016sequence,rennie2017self} such as BLEU~\cite{papineni2002bleu}, ROUGE~\cite{lin2004rouge}, CIDEr~\cite{vedantam2015cider}, SPICE~\cite{anderson2016spice}, or a combination of them~\cite{liu2017improved}. Following this principle, Dai~\etal~\cite{dai2017contrastive} proposed a contrastive loss method to distinguish captions based on their relationship to references, while the approach proposed in~\cite{luo2018discriminability} exploits a reward represented by a weighted combination of the CIDEr score and a discriminability loss. Slightly different is the work proposed by Ren~\etal~\cite{ren2017deep}, which relies on controlling the captioning model by mapping images and sentences into a unified semantic embedding space.

Despite the effectiveness of these training schemes, especially when employed in combination with a CIDEr-based reward, the advent of pre-trained vision-and-language models like CLIP~\cite{radford2021learning} has also shed light on the limitations of the traditional criteria to evaluate caption quality. In fact, while using a CIDEr-based reward can lead to aligning with the style of ground-truth captions, it can also significantly reduce the semantic richness of predicted sentences. Following this premise, our work introduces a novel training strategy, focusing on the complete removal of all reference captions involved in calculating the reward and exploiting the supervision given by a CLIP-based model fine-tuned with additional examples. Along this line, very few approaches~\cite{yu2022multimodal,cho2022fine,dessi2023cross,moratelli2024revisiting} closely aligned with ours refer to the CLIP model to obtain more descriptive captions.

\section{Proposed Method}
\label{sec:method}
\subsection{Preliminaries}
\label{sec:preliminaries}
In this section, we recap the definition of the training protocol typically used in image captioning, of Contrastive Language-Image Pre-training~\cite{radford2021learning}, and of learnable image captioning metrics. Also, we introduce the terminology employed in the rest of the paper.

\tit{Captioning training protocol} Image captioning models are usually trained with a two-stage training approach. The network $f_\theta$ is first pre-trained by encoding an image $I_i$, described through a sequence of $R=(v_1, v_2, ..., v_R)$ visual features, with a time-wise cross-entropy loss in relation to ground-truth sentences $s_{ij} = (w_1, w_2, ..., w_T)$. In the second stage, the network undergoes fine-tuning through a RL strategy aimed at maximizing the CIDEr score~\cite{vedantam2015cider} on the training dataset. 
During the first stage, the model is trained from scratch through a conditioning mechanism, wherein caption generation depends not only on visual features $R$ but also on all previous ground-truth tokens up to time step $t-1$, where $w_t$ is a token belonging to a pre-defined vocabulary. During this phase, $f_\theta$ is optimized using a cross-entropy loss (XE) as follows:
\begin{equation}
\label{eq:xe}
L_{\text{XE}}(\theta) = - \sum_{t=1}^t \log\Bigl(P(w_t | w_{1: t-1}, R)\Bigr).
\end{equation}
The network then operates in an autoregressive manner, generating one token per time step. The model $f_\theta$ outputs a discrete probability distribution, where the token $w_t$ is chosen as the one with the highest probability, determined by preceding tokens. This selection involves passing the final network embeddings through an MLP followed by a softmax function. In the second training stage, at each time step $t$ tokens are sampled from the probability distribution generated by the model at time step $t-1$. Once the entire caption is generated, the CIDEr score is computed as reward to guide a policy-gradient RL update step~\cite{rennie2017self}.

\tit{Contrastive Language-Image Pre-Training (CLIP)}
CLIP~\cite{radford2021learning} represents a state-of-the-art model for the computation of similarities between images and texts. In this context, the computation of matrix similarities and the training of the network through contrastive learning assume a critical role, as it serves as a fundamental step in learning the intrinsic relationships between textual and visual elements, denoted as $T$ and $V$ respectively. The effectiveness of the contrastive method is particularly evident when applied to large-scale datasets. Here, the matrix $T$ is defined as comprising $N_t$ textual instances, each characterized by a $D$-dimensional embedding. Likewise, the visual representation matrix $V$ has a size of $N_v \times D$. To calculate the similarity matrix $S$, the cosine similarity function is adopted. For each textual instance $T_i$ and visual instance $V_j$, the similarity score $S_{ij}$ is computed as follows: $S_{ij} = \text{sim}(T_i, V_j)$, where $\text{sim}(\cdot)$ represents the cosine similarity. This leads to a matrix $S$, with dimensions $N_t \times N_v$, where each element $S_{ij}$ represents the similarity score between the $i$-th textual instance and the $j$-th visual instance.

\begin{figure*}[t]
    \centering
    \includegraphics[width=\linewidth]{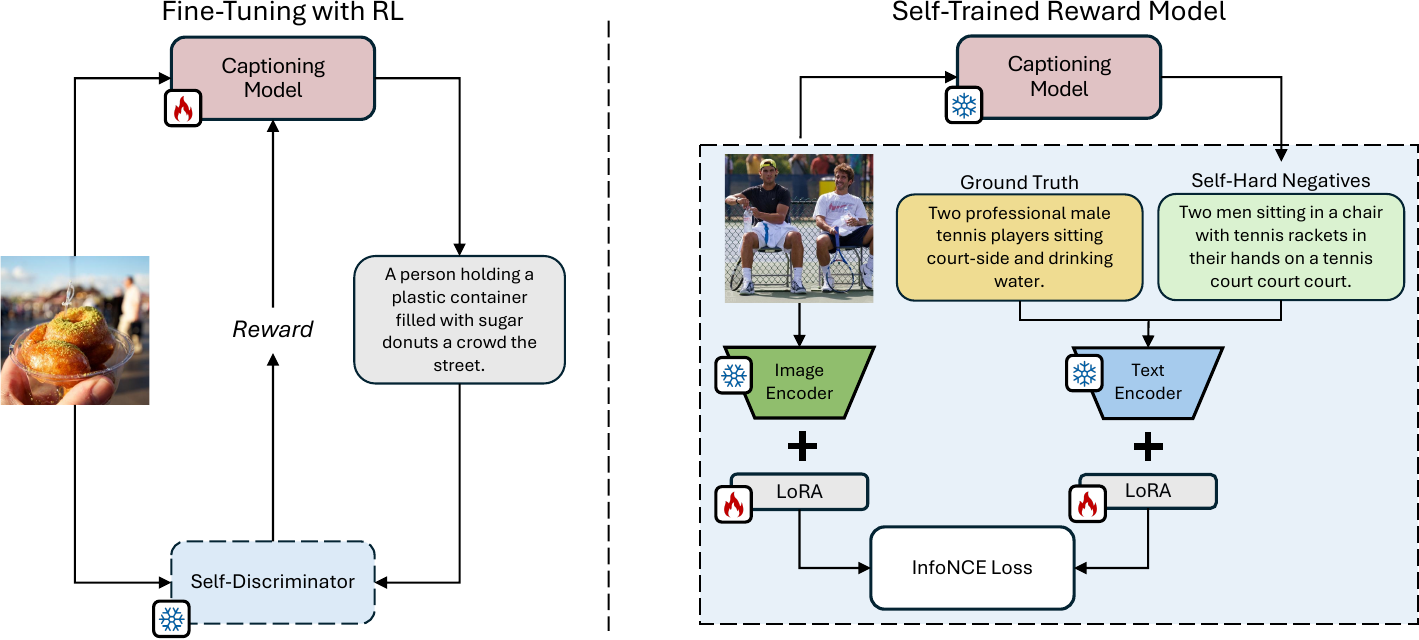}
    \vspace{-0.6cm}
    \caption{Overview of our approach. On the left, the training strategy of the captioner model is shown. The model acts as an agent providing rewards from a discriminator obtained with textual negatives directly derived from the model itself (right).}
    \label{fig:first_page}
    \vspace{-0.35cm}
\end{figure*}

\tit{Learnable captioning metrics from human feedback} 
A recent yet underexplored research direction involves leveraging a model trained with language-image pre-training as an image captioning metric, given its robust alignment capabilities between visual and textual domains. Following~\cite{hessel2021clipscore}, the evaluation score of a caption $s'_i$ can be computed with a cosine similarity $\text{sim}(I_i, s'_i)$ between the visual embedding of the input image and the generated caption. In particular, in~\cite{hessel2021clipscore} a score proportional to the ReLU of the predicted similarity is employed. Additionally, to confine the score within the range of $\left[0,1\right]$  for convenience, the final result is scaled by a multiplicative factor denoted as $w$:
\begin{equation}
    \text{Score}(I_i, s'_i) = w \cdot \text{ReLU}(\text{sim}(I_i, s'_i)).
\end{equation}
One of the most commonly used learnable scores is CLIP-S~\cite{mokady2021clipcap}, where the underlying architecture was pre-trained on 400M noisy (image, text) pairs sourced from the internet. Despite demonstrating better alignment with human judgment compared to traditional captioning metrics (\eg~BLEU, METEOR, CIDEr), which rely on reference captions, the use of noisy data during training leads to significant performance degradation when this score is used to directly optimize a captioning model, resulting in disparities between the score and the overall quality of captions. To mitigate this, a recent approach termed PAC-S~\cite{sarto2023positive} involves fine-tuning the model on cleaned data, thereby enhancing correlation with human evaluations. Specifically, PAC-S score is trained using a similarity matrix constructed from human-curated captions and machine-generated ones. Nevertheless, although these two metrics appear to yield improved correlation with humans, they tend to favor longer texts that are semantically rich yet grammatically flawed over shorter yet grammatically correct captions.

\subsection{Self-Trained Reward Model}
The SCST approach outlined in Sec.~\ref{sec:preliminaries} has proven to be effective in increasing the quality of description with respect to a single XE training stage. However, it also tends to bias the model towards the ``average'' caption that reflects the most general mode contained in the training set~\cite{chen2022learning}. This comes with some critical disadvantages, including reduced descriptiveness, semantic richness, and discriminative power of the generated captions. What is more, one could argue that employing the CIDEr metric as a reward is an obsolete choice, as it achieves a low correlation with human judgments in comparison with recent alternatives.

Following this intuition, in this paper we propose a novel training scheme which is based on a self-supervised reward. In our approach, the classical CIDEr reward is replaced by a learnable language-image discriminator $\mathcal{D}_r$, which takes the form of a language-image model. Following the REINFORCE algorithm, the expected gradient of the reward function can be computed as
\begin{equation}
    \nabla_\theta L_{\text{SCST}}(I_i, s'_i, \theta) = (\mathcal{D}_r(I_i, s'_i) - b) \nabla_\theta \log f_\theta(s'_i),
\end{equation}
where the expected gradient has been approximated using a single Monte-Carlo sample, and $b$ is a baseline employed to reduce the variance of the gradient estimate, which is usually computed as a function of the rewards computed inside a mini-batch. A classical choice when generating multiple descriptions for the same image through beam search is that of computing $b$ as the average reward of all descriptions generated for $I_i$, so that $b=\sum_{j} \mathcal{D}_r(I_i, s'_{ij}) / n$. 

There are three conceptual advantages in replacing an handcrafted captioning metric with a learnable discriminator: (i) contrarily to a standard metric, $\mathcal{D}_r$ is aware of $I_i$ and thus can evaluate image-text alignment by ``looking'' at the image; (ii) being not handcrafted, $\mathcal{D}_r$ can be trained to mimic an evaluation behavior of choice, and does not depend on the annotation style; (iii) $\mathcal{D}_r$ is not limited to work on semantic domains on which ground-truth captions are available.

In this regard, a straightforward choice for $\mathcal{D}_r$ would be that of employing a pre-trained CLIP model based, which also has a large semantic coverage, as explored in~\cite{cho2022fine}. However, when employing learnable rewards, we observed a significant decrease of performance on reference-based metrics, which nonetheless serve as crucial benchmarks for assessing caption quality. Moreover, it is well known that CLIP-based architectures, if not properly fine-tuned, tend to focus heavily on the semantics of the caption, strongly neglecting its grammatical aspect, which is one of the most important aspects of image captioning. From a pragmatic perspective, several works have analyzed the embedding space of CLIP and consistently find that it excels in aligning object categories with images using a bag-of-words approach. This results in robustness against word swapping, rather than mere repetition of identical concepts. Therefore, we introduce a novel fine-tuning methodology grounded in self-supervised learning, which comprises two distinct stages: (i) refinement of CLIP through fine-tuning conditioned on self hard-negatives sourced from the model itself post fine-tuning with CLIP-S and PAC-S; (ii) fine-tuning of the pre-trained model employing our self-discriminator as a reward model.

\subsection{Fine-tuning of Self-Discriminator}
\begin{figure*}[t]
    \centering
    \includegraphics[width=\linewidth]{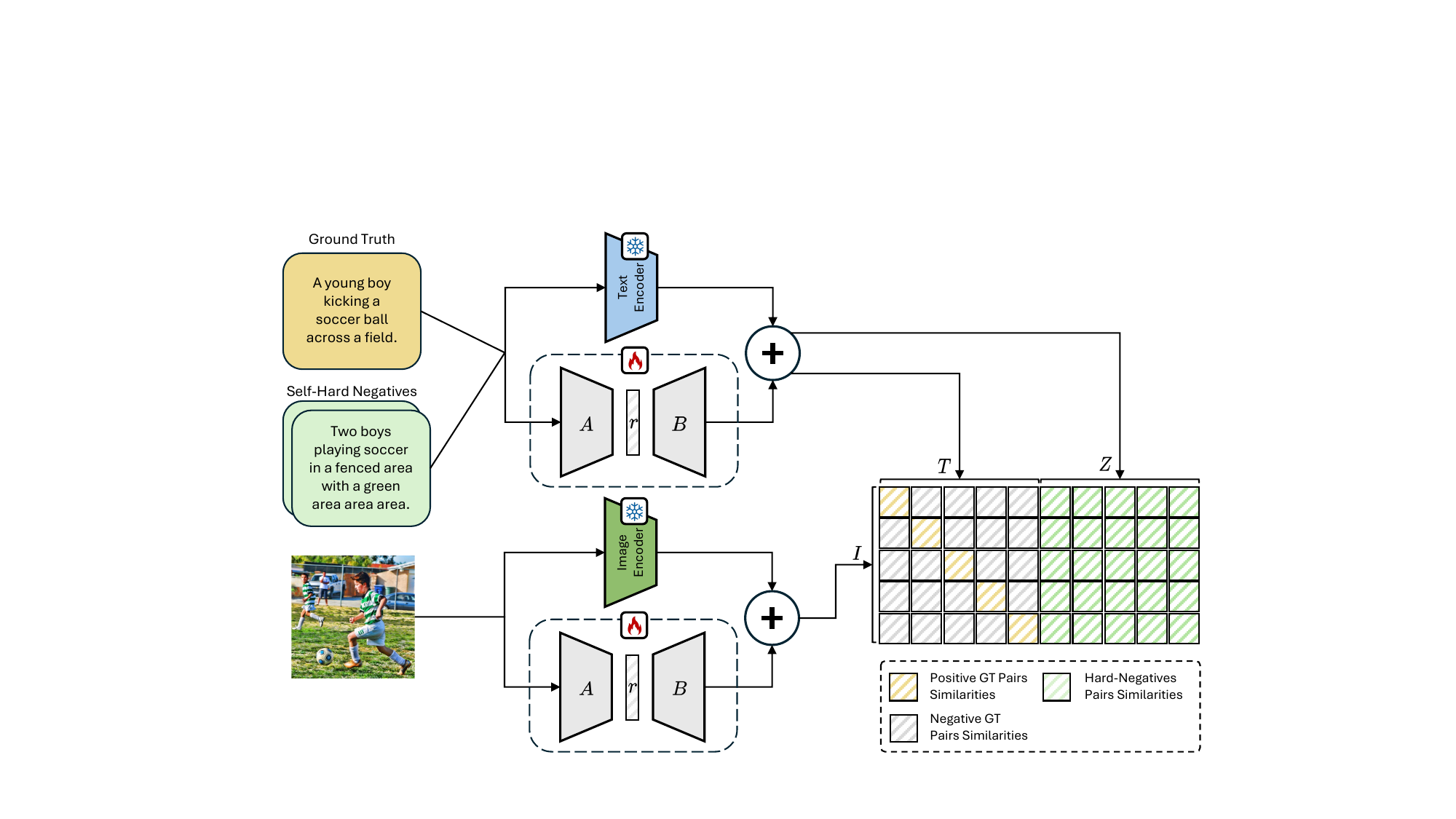}
    \vspace{-0.6cm}
    \caption{Overview of our self-discriminator approach, in which both CLIP encoders are fine-tuned with low-rank adaptation (LoRA) using additional textual negatives.}
    \vspace{-0.35cm}
    \label{fig:clip}
\end{figure*}

As mentioned above, the first stage involves refining the CLIP-based discriminator $\mathcal{D}_r$ through generation-aware mining of hard-negatives. Initially, we employ captioner models trained with CLIP-based rewards to generate these negative instances, which are then exploited to fine-tune CLIP. This process aims to condition CLIP against enforcing alignment styles particularly unsuitable for image captioning. Specifically, through fine-tuning, the goal is to modify the noisy embedding space of CLIP based on the errors obtained from the captioning model. When CLIP is employed in SCST, it results in a meager grammatical reward, despite its strong semantic robustness. For this purpose, we have generated two distinct types of negatives for each sample (\ie~$Z_i = \{{Z_i}^1,{Z_i}^2\}$) derived from the fine-tuned captioner using SCST with rewards based on CLIP-S and PAC-S in their reference-based versions, respectively. This choice allows the model to learn not only to better align the embedding space but also to provide self-supervised reward and thus learn from its own mistakes.

To fine-tune the CLIP-based discriminator $\mathcal{D}_r$, we propose a simple modification to the CLIP objective (see Figure~\ref{fig:clip}). In particular, given a batch of $N$ images $\mathcal{I}=\{I_1, ..., I_N\}$ and $N$ captions $\mathcal{T}=\{T_1, ..., T_N\}$, we concatenate the textual negatives in such a way as to obtain $\bar{\mathcal{T}}={\{T_1, ..., T_N, {Z_1}^1,{Z_1}^2, ..., {Z_N}^1,{Z_N}^2\}}$. Next, we compute the similarity matrix $S \in \mathbb{R}^{N \times 3N}$. Here, the row-wise and column-wise cross-entropy losses are computed as in CLIP, with the difference that we do not compute the loss for the negative captions column-wise (as there is no matching image for a negative caption). To reduce the number of trainable parameters and save memory, we employ low-rank adaptation (LoRA)~\cite{hu2021lora} during the fine-tuning phase of our CLIP-based discriminator, on all layers of both visual and textual encoders. 

\subsection{Training strategy}
Once the fine-tuning of the discriminator is completed, it is employed as a reward signal to fine-tune the captioner through SCST. Our fine-tuned discriminator $\mathcal{D}_r$ is capable of providing feedback not only on semantics but it is also sensitive to grammar and syntax. Finally, the reward perceived by our agent is conditioned not only on the generated text but also on the input image and implicitly on the errors that our model would have generated without any correction and modification of the embedding space.

\section{Experimental Evaluation}
\label{sec:experiments}
\subsection{Datasets and Evaluation Protocol}
We train our model on the COCO dataset~\cite{lin2014microsoft} which contains around 120k images each associated to five different captions, using the splits defined in~\cite{karpathy2015deep} where 5,000 images are used for validation, another 5,000 for testing, and the remainder for training. We then evaluate the effectiveness of our solution on the COCO test set and on the validation set of different image captioning datasets, namely nocaps~\cite{agrawal2019nocaps}, VizWiz~\cite{gurari2020captioning}, and CC3M~\cite{sharma2018conceptual}. 

To evaluate our results, we employ both standard captioning metrics, such as BLEU~\cite{papineni2002bleu}, METEOR~\cite{banerjee2005meteor}, ROUGE~\cite{lin2004rouge}, CIDEr~\cite{vedantam2015cider}, and SPICE~\cite{spice2016}, and more recent learning-based scores like CLIP-Score~\cite{hessel2021clipscore} and PAC-Score~\cite{sarto2023positive} in their reference-free and reference-based versions. In addition, we employ a novel measure to evaluate the grammatical correctness of the generated captions. Specifically, we define Rep-$n$ with $n=1,2,3,4$ as the average number of $n$-grams which are repeated in the generated captions.

\subsection{Implementation Details}
\tinytit{CLIP fine-tuning} Regarding the fine-tuning of CLIP, we use ViT-B/32 as backbone for encoding both images and textual sentences, leveraging the original OpenAI implementation\footnote{\url{https://github.com/openai/CLIP}}. As positive examples, we exploit image-caption pairs from the COCO dataset. We use AdamW~\cite{loshchilov2017decoupled} as optimizer with a learning rate set to $1\cdot 10^{-4}$ and a batch size of 256. Additionally, to reduce the number of trainable parameters and make fine-tuning more efficient, we employ LoRA~\cite{hu2021lora} with a rank equal to 8.

\tit{Architecture}
As our captioning model, we employ a standard encoder-decoder Transformer with 3 layers in both encoder and decoder, a hidden size of 512, and 8 attention heads. To encode input images, we use different CLIP-based backbones, such as RN50, ViT-B/32, and ViT-L/14. To implement our model, we employ the Hugging Face library~\cite{wolf-etal-2020-transformers}.

\tit{Training details}
We first pre-train the model with the classical cross-entropy loss for sentence generation. Next, we optimize our model using different rewards based on unsupervised and supervised metrics (\ie~our \ours strategy, both CLIP-Score~\cite{hessel2021clipscore} and PAC-Score~\cite{sarto2023positive}, and the CIDEr score). During cross-entropy pre-training, we train our network with the Adam optimizer~\cite{kingma2015adam}, a batch size of 1,024, and for up to 20,000 steps. During this phase, we linearly warmup for 1,000 steps, then keep a constant learning rate of $2.5\cdot 10^{-4}$ until 10,000 steps, then sub-linearly decrease until 15,000 steps to $10^{-5}$ and keep the value constant until the end of the training. For the second stage, we further optimize our model with $1\cdot 10^{-6}$ as learning rate using a batch size of 32. During caption generation, we employ a beam size equal to 5.

\subsection{Experimental Results}

\begin{table}[t]
  \centering
    \caption{Comparison between different reward signals in terms of supervised, unsupervised, and grammar-based metrics. Results are reported on the COCO test set.}
  \label{tab:main_table}
  \setlength{\tabcolsep}{.2em}
  \resizebox{\linewidth}{!}{
  \begin{tabular}{lc c ccccccc c cc c cccc}
    \toprule
    & & & \multicolumn{7}{c}{\textbf{Supervised} $\uparrow$} & & \multicolumn{2}{c}{\textbf{Unsupervised} $\uparrow$} & & \multicolumn{4}{c}{\textbf{Grammar} $\downarrow$} \\
    \cmidrule{4-10} \cmidrule{12-13} \cmidrule{15-18}
    \textbf{Backbone} & \textbf{Reward} & & B-4 & M & R & C & S & RefCLIP-S & RefPAC-S & & CLIP-S & PAC-S & & Rep-1 & Rep-2 & Rep-3 & Rep-4 \\
    \midrule
    & - & & 32.8 & 28.1 & 55.0 & 109.8 & 20.3 & 0.796 & 0.853 & & 0.743 & 0.817 & & 1.516 & 0.108 & 0.022 & 0.009 \\
    & CIDEr & & 39.7 & 29.2 & 58.3 & 126.8 & 21.2 & 0.797 & 0.855 & & 0.739 & 0.817 & & 1.384 & 0.05 & 0.008 & 0.005 \\
    \cmidrule{2-18}
    & CLIP-S & & 14.3 & 24.7 & 34.9 & 3.1 & 21.2 & 0.765 & 0.830 & & 0.804 & 0.837 & & 11.762 & 5.168 & 2.809 & 1.518 \\
    & PAC-S & & 18.5 & 26.5 & 42.2 & 32.2 & 21.7 & 0.785 & 0.849 & & 0.799 & \textbf{0.860} & & 5.453 & 1.588 & 0.645 & 0.288 \\
    & CLIP-S~\cite{cho2022fine} & & 6.3 & 19.7 & 29.5 & 11.2 & 12.3 & 0.786 & 0.823 & & \textbf{0.843} & 0.837 & & 5.619 & 1.541 & 0.466 & 0.151 \\
    & CLIP-S+Gr~\cite{cho2022fine} & & 16.9 & 25.9 & 45.6 & 71.2 & 19.6 & \textbf{0.792} & 0.849 & & 0.779 & 0.839 & & \textbf{1.536} & \textbf{0.097} & \textbf{0.015 }& \textbf{0.003} \\
    \rowcolor{Gray}
    \cellcolor{white} \multirow{-7}{*}{\textbf{RN50}} & \textbf{\ours} & & \textbf{20.8} & \textbf{26.8} & \textbf{48.2} & \textbf{72.0} & \textbf{21.8} & \textbf{0.792} & \textbf{0.851} & & 0.780 & 0.844 & & 2.706 & 0.495 & 0.153 & 0.049 \\
    \midrule
    & - & & 33.1 & 28.2 & 55.4 & 112.4 & 20.5 & 0.804 & 0.861 & & 0.755  & 0.830 & & 1.468 & 0.091 & 0.017 & 0.005 \\
    & CIDEr & & 39.4 & 29.5 & 58.3 & 129.0 & 22.2 & 0.809 & 0.866 & & 0.757 & 0.833 & & 1.360 & 0.055 & 0.006 & 0.001 \\
    \cmidrule{2-18}
    & CLIP-S & & 11.4 & 23.1 & 31.2 & 1.1 & 18.5 & 0.778 & 0.830 & & \textbf{0.851} & 0.846 & & 11.166 & 3.566 & 1.232 & 0.395 \\
    & PAC-S & & 20.3 & 27.1 & 44.1 & 40.7 & 22.4 & 0.796 & 0.858 & & 0.810  & \textbf{0.870} & & 5.078 & 1.443 & 0.584 & 0.260 \\
    \rowcolor{Gray}
    \cellcolor{white}\multirow{-5}{*}{\textbf{ViT-B/32}} & \textbf{\ours} & & \textbf{23.6} & \textbf{27.3} & \textbf{49.3} & \textbf{81.4} & \textbf{22.9} & \textbf{0.808} & \textbf{0.862} & & 0.800 & 0.861 & & \textbf{2.626} & \textbf{0.483} & \textbf{0.156} & \textbf{0.063} \\
    \midrule
    & - & & 37.3 & 30.4 & 58 1& 126.6 & 23.3 & 0.811 & 0.868 & & 0.758  & 0.831 & & 1.402 & 0.062 & 0.007 & 0.002 \\
    & CIDEr & & 43.6 & 30.8 & 61.0 & 143.3 & 23.2 & 0.809 & 0.866 & & 0.750 & 0.826 & & 0.239 & 0.498 & 0.616 & 0.349 \\
    \cmidrule{2-18}
    & CLIP-S & & 10.2 & 23.0 & 30.3 & 1.1 & 15.3 & 0.793 & 0.827 & & \textbf{0.865} & 0.834 & & 8.788 & 2.113 & 0.716 & 0.248 \\
    & PAC-S & & 22.3 & 28.4 & 46.2 & 51.1 & 24.6 & 0.801 & 0.861 & & 0.805 & \textbf{0.862} & & 4.612 & 1.199 & 0.479 & 0.206 \\
    \rowcolor{Gray}
    \cellcolor{white}\multirow{-5}{*}{\textbf{ViT-L/14}} & \textbf{\ours} & & \textbf{22.6} & \textbf{28.4} &\textbf{50.2} & \textbf{82.7} & \textbf{24.7} & \textbf{0.809} &\textbf{0.864} & & 0.787 & 0.853 & & \textbf{2.216} & \textbf{0.376} & \textbf{0.118} & \textbf{0.039} \\
    \bottomrule
  \end{tabular}
}
\vspace{-0.2cm}
\end{table}

\tinytit{Results on COCO test set} We start by comparing our solution against other CLIP-based rewards (\ie~CLIP-S and PAC-S) using different visual backbones to encode input images. Results are reported in Table~\ref{tab:main_table} in terms of supervised, unsupervised, and grammar-based metrics. For completeness, we also include the results of the model trained after cross-entropy loss and using a standard CIDEr score as reward. In all experiments, we employ the same Transformer-based architecture with three layers in both the encoder and decoder. Regarding a comparison with previous works, it is important to note that the only work within the same settings is proposed by Cho~\etal~\cite{cho2022fine} which however only adopts CLIP RN50 backbone as visual encoder. Specifically, two variants both optimized using CLIP-S are proposed, where the former only employs CLIP-S as reward while the latter combines CLIP-S with a grammar-based reward.

From the results, we can notice that adopting a reward relying on CLIP-based models significantly alters the performance of the model, leading to word repetitions and a lack of logical or grammatical structure within the caption. Indeed, within a few steps, the model appears to hack the metric by finding alternative ways to boost the semantics and consequently the value of the metric itself (\ie~CLIP-S or PAC-S), completely disregarding the syntactic structure of the caption. In particular, considering the results of our proposal (\ie~\ours) with ViT-B/32 as visual backbone, it can be seen that our reward strategy can significantly improve the results on standard supervised metrics (\eg~81.4 CIDEr points compared to 40.7 and 1.1 achieved with PAC-S and CLIP-S rewards respectively). This demonstrates the effectiveness of \ours in better preserving the coherence of the predicted caption with the image and the ability to generate ``human-like'' and thus structurally correct captions. As expected, directly optimizing a specific metric leads to the best results on that metric, as showed by the results of the models trained with CLIP-S or PAC-S as reward. Nonetheless, this is not confirmed on the reference-based versions of CLIP-S and PAC-S for which \ours achieves the best performance according to all employed backbones, further confirming a better correlation with human-written captions.

\begin{table}[t]
  \centering
  \caption{Descriptiveness analysis of generated captions in terms of unsupervised scores and retrieval-based metrics. Results are reported on the COCO test set.}
  \label{tab:coco_retrieval}
  \setlength{\tabcolsep}{.38em}
  \resizebox{0.7\linewidth}{!}{
  \begin{tabular}{lc c cc c cccc}
    \toprule
    & & & \multicolumn{2}{c}{\textbf{Unsupervised}} & & \multicolumn{4}{c}{\textbf{Recall}} \\
    \cmidrule{4-5} \cmidrule{7-10}
    \textbf{Backbone} & \textbf{Strategy} & & CLIP-S & PAC-S & & R@1 & R@5 & R@10 & MRR \\
    \midrule
    & XE & & 0.743 & 0.817 & & 21.2 & 44.2 & 57.6 & 31.2 \\
    & SCST (CIDEr) & & 0.739 & 0.817 & & 19.8 & 43.4 & 55.7 & 29.8 \\
    \rowcolor{Gray}
    \cellcolor{white} \multirow{-3}{*}{\textbf{RN50}} & \textbf{\ours} & & \textbf{0.780} & \textbf{0.844} & & \textbf{37.7} & \textbf{67.3} & \textbf{78.6} & \textbf{50.3 }\\
    \midrule
    & XE & & 0.755 & 0.830 & & 24.8 & 50.8 & 62.8 & 35.7 \\
    & SCST (CIDEr) & & 0.757 & 0.833 & & 25.7 & 51.7 & 64.4 & 36.7 \\
    \rowcolor{Gray}
    \cellcolor{white} \multirow{-3}{*}{\textbf{ViT-B/32}} & \textbf{\ours} & & \textbf{0.800} & \textbf{0.861} & & \textbf{47.1} & \textbf{74.6} & \textbf{84.9} & \textbf{58.9} \\
    \midrule
    & XE & & 0.758 & 0.831 & & 27.7 & 52.6 & 64.2 & 38.5 \\
    & SCST (CIDEr) & & 0.750 & 0.826 & & 23.9 & 49.8 & 61.6 & 34.9 \\
    \rowcolor{Gray}
    \cellcolor{white} \multirow{-3}{*}{\textbf{ViT-L/14}} & \textbf{\ours} & & \textbf{0.787} & \textbf{0.853} & & \textbf{44.7} & \textbf{71.8} & \textbf{82.6} & \textbf{56.5} \\
    \bottomrule
  \end{tabular}
}
\vspace{-0.2cm}
\end{table}

To further clarify the problems associated with unsupervised metrics when used as rewards, we also report the average number of repeated $n$-grams for each caption (\ie~Rep-$n$ with $n=1,2,3,4$). Notably, \ours significantly reduces the number of repetitions within the generated sentences, decreasing the 1-gram repetitions from 11.166 and 5.078 respectively using CLIP-S and PAC-S to 2.626, always when employing visual features from ViT-B/32. These results are confirmed also considering a larger number of $n$-grams and across all considered visual backbones, further demonstrating the effectiveness of our training strategy in reducing the grammatical incorrectness of captions generated by captioners optimized using standard CLIP-based rewards.

When instead comparing our model with the one proposed in~\cite{cho2022fine} using RN50 visual features, we can notice that the model optimized only with CLIP-S version yields a high value of CLIP-S, while totally degrading the reference-free metrics (\ie~11.2 CIDEr points with respect to 72.0 of \ours) and producing numerous repetitions (\ie~5.619 and 1.541 of Rep-1 and Rep-2 compared to 2.706 and 0.495 of our approach). The scenario is different when considering the second variant, which is optimized with a combination of CLIP-S and a grammar-based reward. Specifically, while \ours still achieves higher results in terms of all supervised metrics, it presents slightly higher values of repetitions. Nevertheless, it is noteworthy that \ours does not exploit any explicit grammatical reward, as it is learned directly within the embedding space of the discriminator itself during the refinement process.

\tit{Analysis on the descriptiveness of generated captions}
To effectively compare the captions generated by \ours with those generated by a captioning model trained with a standard training paradigm (\ie~cross-entropy loss followed by SCST with CIDEr reward), we complement the results shown in Table~\ref{tab:main_table} with retrieval-based metrics reported in Table~\ref{tab:coco_retrieval}. Retrieval-based metrics are generally used to measure the discriminative degree of the generated captions, which is usually a viable strategy to estimate their descriptiveness and semantic richness.

In particular, following recent works~\cite{kornblith2023guiding,chan2023ic}, we measure the quality of generated captions in distinguishing images in a dataset and compute the percentage of the times the image corresponding to each generated caption is retrieved among the first $k$ retrieved items. This is done by ranking the images in terms of CLIP similarity between visual and textual embeddings, using the CLIP ViT-B/32 model, and computing recall at $K$ with $k=1,5,10$. We also compute the mean reciprocal rank (MRR) for each generated caption: higher MRR scores indicate that captions are more discriminative and therefore usually more detailed. Notably, \ours can significantly increase the results obtained with a standard training paradigm (\ie~24.8 and 25.7 achieved by XE and SCST (CIDEr) in terms of R@1 vs. 47.1 achieved by \ours with ViT-B/32), highlighting a higher degree of descriptiveness in generated captions.

\begin{table}[t]
  \centering
    \caption{Ablation study on COCO test set, using different negative textual sentences and CLIP ViT-B/32 as image encoder.}
  \label{tab:ablation}
  \setlength{\tabcolsep}{.25em}
  \resizebox{0.92\linewidth}{!}{
  \begin{tabular}{cccccccccc c ccc}
    \toprule
    \multicolumn{3}{c}{\textbf{Negatives}} & & \multicolumn{7}{c}{\textbf{Supervised}} && \multicolumn{2}{c}{\textbf{Unsupervised}}\\
    \cmidrule{1-3} \cmidrule{5-11} \cmidrule{13-14}
    Manual & CLIP-S & PAC-S & & B-4 & M & R & C & S & RefCLIP-S & RefPAC-S & & CLIP-S & PAC-S  \\
    \midrule
    \checkmark & & & & 19.7 & 27.4 & 44.0 & 41.2 & \textbf{22.3} & 0.799 & 0.856 & & \textbf{0.812} & \textbf{0.865} \\
    & \checkmark & & & 21.6 & \textbf{27.5} & 46.2 & 57.3 & 22.3 & 0.801 & 0.858 & & 0.808 & 0.865 \\
    & & \checkmark & & 23.1 & 27.4 & 48.5 & 78.9 & 21.9 & 0.805 & 0.861 & & 0.803 & 0.864 \\
    \midrule
    \checkmark & & \checkmark & & 21.3 & 27.1 & 47.5 & 70.0 & 21.8 & 0.807 & 0.862 & & 0.798 & 0.861 \\
    \checkmark & \checkmark & \checkmark & & 21.0 & 27.3 & 46.0 & 60.4 & 21.7 & \textbf{0.808} & \textbf{0.862} & & 0.802 & 0.862 \\
    \midrule
    \rowcolor{Gray}
    & \checkmark & \checkmark & & \textbf{23.6} & 27.3 & \textbf{49.3} & \textbf{81.4} & 21.9 & \textbf{0.808} & \textbf{0.862} & & 0.800 & 0.861 \\
    \bottomrule
  \end{tabular}
  }
  \vspace{-0.3cm}
\end{table}

\tit{Ablation study on negative examples}
As mentioned in Sec.~\ref{sec:method}, to compute the reward during the RL-based optimization, we employ a CLIP-based discriminator fine-tuned using a combination of self-generated negative samples obtained by two different captioners, one trained with CLIP-S reward and the other trained with PAC-S reward. In Table~\ref{tab:ablation}, we evaluate the effectiveness of the chosen negative samples. In particular, we consider negative samples generated by a single captioning model (\ie~either trained with CLIP-S or PAC-S) and manually-constructed negative samples, or a combination of them. When generating manual negatives, we consider the failure cases typically produced by a captioner fine-tuned with CLIP-based rewards: (i) premature termination of captions (\eg~``a man playing with a cat in''); (ii) redundancy of the final term (\eg~``a man with an umbrella in the background background background''); and (iii) duplication of concepts within captions (\eg~``a cat in the garden and a cat in the garden''). We therefore manually corrupt COCO captions either manually repeating or removing one or more random words, performing a random swap of two words, or substituting one word with a randomly selected word from the entire vocabulary of the COCO dataset.

As it can be seen, the best results are obtained using a combination of negative samples deriving from the combination of CLIP-S and PAC-S, which achieves significantly higher CIDEr values compared to the manually created negatives (\ie~81.4 vs. 41.2) and all other alternatives. Overall, the use of manual negatives does not prove effective also when used in combination with other considered negative samples, leading to performance degradation on all supervised metrics.

\tit{Out-of-domain evaluation}
To assess the out-of-domain capabilities of our model, we evaluated Self-Cap on three distinct datasets, namely nocaps~\cite{agrawal2019nocaps}, CC3M~\cite{sharma2018conceptual}, and VizWiz~\cite{gurari2020captioning}. While nocaps is specifically tailored for the novel object captioning task encompassing object classes absent in COCO, CC3M and VizWiz respectively comprises images sourced from the web and captured by visually impaired people. Except for captions from CC3M which are automatically generated, all other datasets are composed of manually-curated textual sentences. Table~\ref{tab:out_domain} shows the results obtained using three different visual backbones, comparing our approach with models fine-tuned using CLIP-S and PAC-S rewards. Also in this setting, \ours achieves significantly higher results in terms of standard evaluation metrics, demonstrating the effectiveness and generalization capabilities of our approach even in out-of-domain scenarios.

\begin{table}[t]
  \caption{Out-of-domain performance analysis on nocaps, VizWiz, and CC3M validation sets in terms of supervised and unsupervised metrics.}
  \label{tab:out_domain}
  \vspace{-0.08cm}
  \centering
  \setlength{\tabcolsep}{.18em}
  \resizebox{\linewidth}{!}{
  \begin{tabular}{lc c ccccccc c ccccccc c ccccccc}
    \toprule
    & & & \multicolumn{6}{c}{\textbf{nocaps}} & & \multicolumn{6}{c}{\textbf{VizWiz}} & & \multicolumn{6}{c}{\textbf{CC3M}} \\
    \cmidrule{4-9} \cmidrule{11-16} \cmidrule{18-23}
    \textbf{Backbone} & \textbf{Reward} & & B-4 & R & C & S & CLIP-S & PAC-S & & B-4 & R & C & S & CLIP-S & PAC-S & & B-4 & R & C & S & CLIP-S & PAC-S \\
    \midrule
    & CLIP-S & & 3.7 & 23.2 & 4.6 & 12.9 & 0.738 & 0.799 & & 8.70 & 29.8 & 6.7 & 8.8 & 0.667 & 0.78 & & 1.0 & 13.9 & 4.3 & 6.5 & 0.678 & 0.78 \\
    & PAC-S & & 4.0 & 25.3 & 20.9 & \textbf{14.1} & \textbf{0.741} & \textbf{0.850} & & 9.22 & 31.6 & 13.01 & 10.3 & \textbf{0.688} & \textbf{0.816} & & 0.8 & 12.4 & 5.8 & 6.5 & \textbf{0.699} & \textbf{0.814} \\
    \rowcolor{Gray}
    \cellcolor{white}\multirow{-3}{*}{\textbf{RN50}} & \textbf{\ours} & & \textbf{4.9} & \textbf{27.1} & \textbf{30.4} & 13.9 & 0.737 & 0.844 & & \textbf{10.1} & \textbf{35.4} & \textbf{19.7} & 8.1 & 0.667 & 0.795 & & \textbf{1.2} & \textbf{14.9} & \textbf{15.9} & \textbf{7.7} & 0.686 & 0.798 \\
    \midrule
    & CLIP-S & & 4.0 & 27.1 & 9.8 & 13.2 & \textbf{0.754} & 0.810 & & 5.5 & 23.8 & 1.3 & 8.5 & \textbf{0.737} & 0.814 & & 0.8 & 11.4 & 0.6 & 6.0 & \textbf{0.718} & 0.784 \\
    & PAC-S & & 5.2 & 28.5 & 35.7 & \textbf{16.2} & 0.750 & \textbf{0.854} & & 11.0 & 34.3 & 20.1 & \textbf{9.8} & 0.715 & \textbf{0.837} & & 1.2 & 14.1 & 9.8 & 7.6 & 0.698 & \textbf{0.809} \\
    \rowcolor{Gray}
    \cellcolor{white}\multirow{-3}{*}{\textbf{ViT-B/32}} & \textbf{\ours} & & \textbf{6.2} & \textbf{29.8} & \textbf{46.3} & 16.0 & 0.751 & \textbf{0.854} & & \textbf{13.0} & \textbf{37.8} & \textbf{27.0} & 9.1 & 0.702 & 0.828 & & \textbf{1.3} & \textbf{15.2} & \textbf{19.4} &\textbf{8.5} & 0.688 & 0.803 \\
    \midrule
    & CLIP-S & & 5.2 & 28.9 & 10.2 & 17.3 & \textbf{0.750} & 0.819 & & 4.1 & 21.8 & 1.2 & 7.0 & \textbf{0.766} & 0.775 & & 0.6 & 10.2 & 0.6 & 4.4 & \textbf{0.747} & 0.765 \\
    & PAC-S & & 5.7 & 30.0 & 44.8 & \textbf{18.1} & 0.746 & \textbf{0.850} & & 11.2 & 36.0 & 26.8 & \textbf{12.2} & 0.701 & \textbf{0.820} & & 1.4 & 15.1 & 13.2 & 8.6 & 0.701 & \textbf{0.811} \\
    \rowcolor{Gray}
    \cellcolor{white}\multirow{-3}{*}{\textbf{ViT-L/14}} & \textbf{\ours} & & \textbf{6.9} & \textbf{31.3} & \textbf{62.8} & \textbf{18.1} & 0.742 & 0.839 & & \textbf{11.4} & \textbf{37.4} & \textbf{28.5} & 10.2 & 0.690 & 0.809 & & \textbf{1.6} & \textbf{16.7} & \textbf{21.9} & \textbf{9.6} & 0.696 & 0.809 \\
    \bottomrule
  \end{tabular}
}
\vspace{-0.3cm}
\end{table}

\subsection{Qualitative Analysis}
\begin{figure*}[t]
    \begin{minipage}{0.165\linewidth}
        \includegraphics[width=0.97\linewidth]{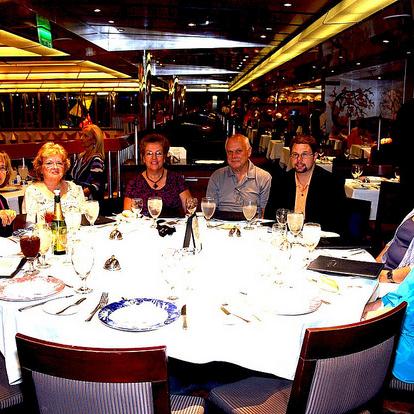}
        \end{minipage}
        \begin{minipage}{0.32\linewidth}
        \scriptsize{
        \textbf{PAC-S:} A group of people sitting at a dinner table with a wine glass in the background of a boat setting of wine in the background of a restaurant.\\
        \textbf{\ours (Ours):} A group of men and women sitting around a dinner table at restaurant.
        }
    \end{minipage}
    \hspace{0.02cm}
    \begin{minipage}{0.165\linewidth}
        \includegraphics[width=0.97\linewidth]{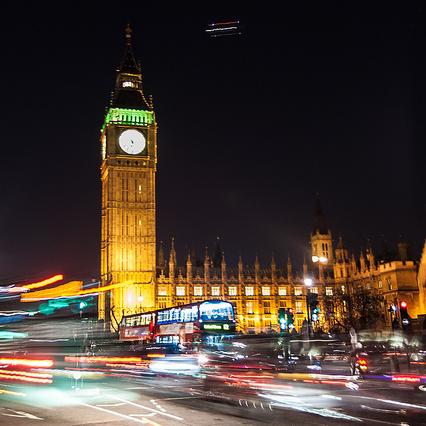}
        \end{minipage}
        \begin{minipage}{0.32\linewidth}
        \scriptsize{
        \textbf{PAC-S:} The big ben clock tower towering over the city of London at night at night time with cars driving past it at night.\\
        \textbf{\ours (Ours):} The big ben clock tower towering over the city of London at night.
        }
    \end{minipage}
    
    \vspace{0.1cm}

    \begin{minipage}{0.165\linewidth}
        \includegraphics[width=0.97\linewidth]{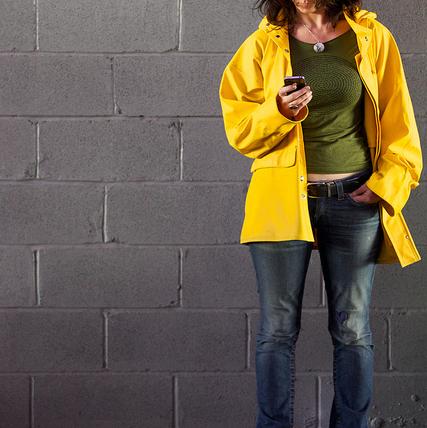}
        \end{minipage}
        \begin{minipage}{0.32\linewidth}
        \scriptsize{
        \textbf{PAC-S:} A woman in a yellow raincoat checking her cell phone against a grey wall with yellow raincoat in the background.\\
        \textbf{\ours (Ours):} A woman in a yellow jacket looking at her cell phone against a brick wall.
        }
    \end{minipage}
    \hspace{0.02cm}
    \begin{minipage}{0.165\linewidth}
        \includegraphics[width=0.97\linewidth]{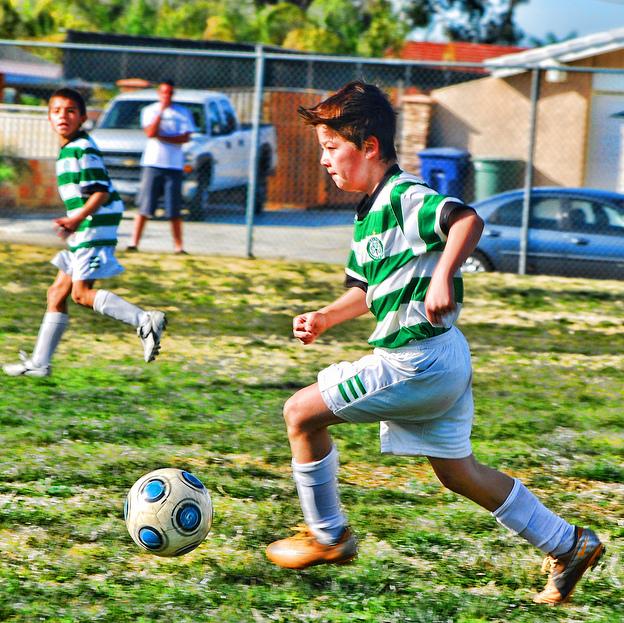}
        \end{minipage}
        \begin{minipage}{0.32\linewidth}
        \scriptsize{
        \textbf{PAC-S:} Two boys playing soccer in a fenced area with a green soccer ball in the background of a home area setting.\\
        \textbf{\ours (Ours):} A young boy kicking a soccer ball in a field with other players.
        }
    \end{minipage}

    \vspace{0.1cm}

    \begin{minipage}{0.165\linewidth}
    \includegraphics[width=0.97\linewidth]{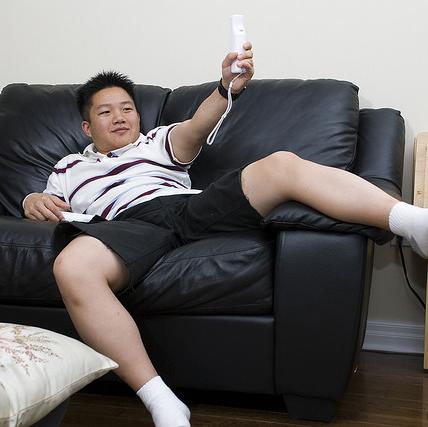}
        \end{minipage}
        \begin{minipage}{0.32\linewidth}
        \scriptsize{
        \textbf{PAC-S:} A young man sitting on a couch holding a wii remote control in his hand while playing video games in the living room area area.\\
        \textbf{\ours (Ours):} A young man laying in a black leather couch holding a wii remote.
        }
    \end{minipage}
    \hspace{0.02cm}
    \begin{minipage}{0.165\linewidth}
        \includegraphics[width=0.97\linewidth]{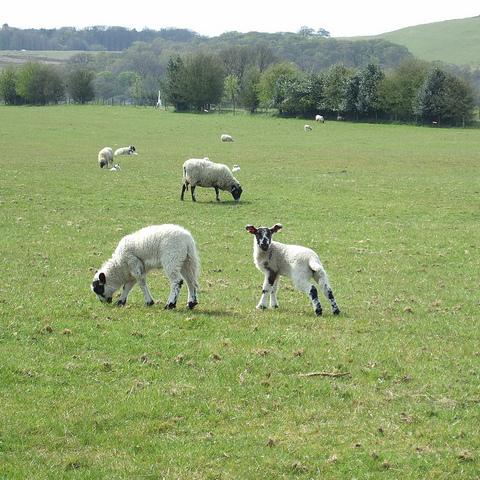}
        \end{minipage}
        \begin{minipage}{0.32\linewidth}
        \scriptsize{
        \textbf{PAC-S:} A herd of sheep grazing on a lush green field with a baby sheep grazing in the background of the background area area of a country.\\
        \textbf{\ours (Ours):} Three sheep are grazing in a grassy field and one is looking at the camera.
        }
    \end{minipage}
\vspace{-.15cm}
\caption{Qualitative results on COCO sample images, comparing \ours with a model trained using PAC-S as reward.}
\label{fig:qualitatives}
\vspace{-.2cm}
\end{figure*}

To validate the quality of captions generated by our approach, Figure~\ref{fig:qualitatives} shows some qualitative samples from the COCO test set. In this case, we compare captions generated by \ours with those generated by a captioning model trained with PAC-S reward. As it can be seen, \ours can generate more descriptive and complex captions while minimizing repetitions and grammatical errors often encountered when combining SCST with CLIP-based rewards.

\section{Conclusion}
\label{sec:conclusion}
We present \ours, a novel fine-tuning method for image captioning which entails a two-phase training procedure. It leverages a discriminator to provide feedback by learning directly from the errors of the captioner. In a setting utilizing a CLIP-based reward, the proposed solution demonstrates state-of-the-art performance in supervised metrics. Additionally, we showcase the out-of-domain capabilities of our approach on three different datasets. \ours generates captions that are not only more complex and semantically richer but also yield superior grammatical accuracy compared to competitors.

\subsubsection{Acknowledgements} We acknowledge the CINECA award under the ISCRA initiative, for the availability of high-performance computing resources and support. This work has been conducted under a research grant co-funded by Altilia s.r.l. and supported by the PRIN 2022 project ``MUSMA'' (CUP G53D23002930006) and by the PRIN 2022-PNRR project ``MUCES'' (CUP E53D23016290001), both funded by EU - Next-Generation EU - M4 C2 I1.1.

%
%
%
\bibliographystyle{splncs04}
\bibliography{bibliography}
\end{document}